\documentclass{article}

\PassOptionsToPackage{numbers, compress}{natbib}

\usepackage[final]{neurips_2025}

\usepackage[utf8]{inputenc} %
\usepackage[T1]{fontenc}    %
\usepackage{hyperref}       %
\usepackage{url}            %
\usepackage{booktabs}       %
\usepackage{amsfonts}       %
\usepackage{nicefrac}       %
\usepackage{microtype}      %
\usepackage[dvipsnames]{xcolor}
\usepackage{xspace}

\usepackage{amssymb}%
\usepackage{pifont}%
\usepackage{paralist}
\usepackage{graphicx}
\usepackage{tabularx}
\usepackage{amsmath}
\usepackage{cleveref}       %
\usepackage{wrapfig}
\usepackage{caption}
\usepackage{subcaption}
\usepackage{colortbl}
\usepackage{tikz}
\usetikzlibrary{positioning}
\usetikzlibrary{calc}

\newcommand{\cmark}{\ding{51}}%
\newcommand{\xmark}{\ding{55}}%

\newcommand{\Ours}{\textbf{Ours}}
\newcommand{\ours}{\textbf{ours}}
\newcommand{\PAR}[1]{\textbf{#1.}\xspace}

\newcommand{\ie}{\textit{ie.}\xspace}

\newcommand{\suppmat}{\textit{supp. mat.}\xspace}

\newcommand{\ifpublic}[1]{}

\crefname{figure}{Fig.}{Figs.}
\Crefname{figure}{Fig.}{Figs.}
\crefname{equation}{Eq.}{Eqs.}
\Crefname{equation}{Eq.}{Eqs.}
\crefname{table}{Tab.}{Tabs.}
\Crefname{table}{Tab.}{Tabs.}
\renewcommand{\ifpublic}[1]{#1}

\title{LODGE: Level-of-Detail Large-Scale Gaussian Splatting with Efficient Rendering}
\author{
Jonas Kulhanek$^{1,4}$, 
Marie-Julie Rakotosaona$^{1}$, 
Fabian Manhardt$^{1}$, 
Christina Tsalicoglou$^{1}$,\\
\textbf{
Michael Niemeyer$^{1}$, 
Torsten Sattler$^{5}$, 
Songyou Peng$^{2}$, 
Federico Tombari$^{1,3}$
}\\
$^{1}$ Google, \quad
$^{2}$ Google DeepMind, \quad
$^{3}$ Technical University of Munich, \\
$^{4}$ Czech Technical University in Prague, Faculty of Electrical Engineering, \\
$^{5}$ Czech Technical University in Prague, Czech Institute of Informatics, Robotics and Cybernetics
}

\definecolor{mycitecolor}{HTML}{195a66}
\hypersetup{
  citecolor  = mycitecolor,
  colorlinks = true,
}
\AtBeginDocument{%
  \urlstyle{tt}
}

\newread\imgstream
\immediate\openin\imgstream=imagedata.in
\makeatletter
\def\new@kvginclip#1{}
\def\new@kvgintrim#1{}
\let\old@kvginclip\KV@Gin@clip
\let\old@kvgintrim\KV@Gin@trim
\let\oldincludegraphics\includegraphics
\providecommand{\includegraphics}{}
\renewcommand{\includegraphics}[2][]{%
  \immediate\read\imgstream to \src
  \immediate\read\imgstream to \removecrop
  \ifnum\removecrop=1
      \let\KV@Gin@clip\new@kvginclip
      \let\KV@Gin@trim\new@kvgintrim
  \fi
  \oldincludegraphics[#1]{\src}%
  \let\KV@Gin@clip\old@kvginclip
  \let\KV@Gin@trim\old@kvgintrim}
\makeatother

\begin{document}

\maketitle

\ifpublic{
\vspace{-10mm}
\begin{center}
\url{https://lodge-gs.github.io/}
\end{center}
\vspace{6mm}
}

\begin{figure}[h]\vspace{-6mm}\centering
\includegraphics[width=\textwidth]{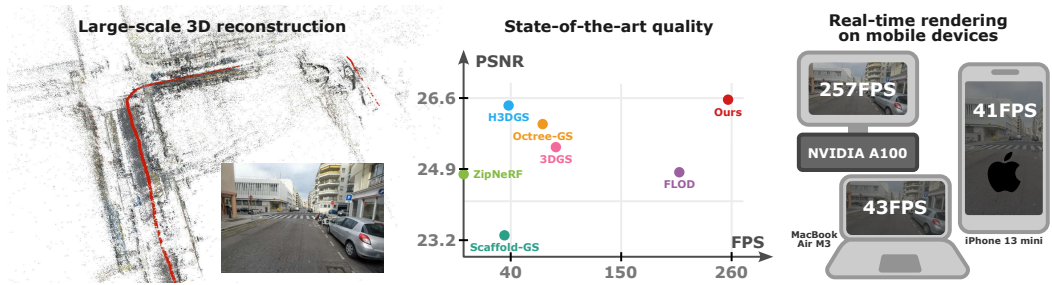}
\vspace{-0.5em}
\caption{\textbf{LODGE.} Applied to large-scale 3D scenes, LODGE achieves outstanding quality while maintaining superior rendering speeds. Furthermore, it enables real-time rendering on mobile devices.
\label{fig:teaser}%
}
\end{figure}

\begin{abstract}
In this work, we present a novel level-of-detail (LOD) method for 3D Gaussian Splatting that enables real-time rendering of large-scale scenes on memory-constrained devices. Our approach introduces a hierarchical LOD representation that iteratively selects optimal subsets of Gaussians based on camera distance, thus largely reducing both rendering time and GPU memory usage. We construct each LOD level by applying a depth-aware 3D smoothing filter, followed by importance-based pruning and fine-tuning to maintain visual fidelity. To further reduce memory overhead, we partition the scene into spatial chunks and dynamically load only relevant Gaussians during rendering, employing an opacity-blending mechanism to avoid visual artifacts at chunk boundaries. Our method achieves state-of-the-art performance on both outdoor (Hierarchical 3DGS) and indoor (Zip-NeRF) datasets, delivering high-quality renderings with reduced latency and memory requirements.
\end{abstract}

\section{Introduction}
Novel view synthesis is a central area of research in computer vision that enables applications in AR/VR, gaming, interactive maps, and others. The field has recently received a lot of attention with the advent of Neural Radiance Fields (NeRFs)~\cite{mildenhall2021nerf} and 3D Gaussian Splatting (3DGS)~\cite{kerbl20233dgs} — the latter pushing the range of applications further as it enables real-time rendering. With the rising popularity of NeRFs and 3DGS, there is an increasing interest in applying such methods to ever larger and more complex scenes~\cite{rematas2022urf,kerbl2024h3dgs,fischer2024dynamic}. However, standard methods do not scale well to such large-scale environments~\cite{kerbl2024h3dgs,liu2024citygs}. The core issue lies in the representation: to capture fine details, a high number of Gaussians is required. Consequently, even distant regions of the scene are populated with dense Gaussians representing fine-grained geometry, \ie, elements that contribute little to the final rendered image. This leads to significant inefficiencies during rendering, as many far-away Gaussians are processed despite having minimal or no visible impact. Furthermore, memory limitations pose an additional challenge: not all Gaussians can fit into GPU memory simultaneously, which is particularly problematic for mobile or low-end devices where memory is severely constrained.

In computer graphics this problem has been extensively studied and — in the context of mesh-based rendering — effectively addressed using level-of-detail (LOD) strategies. These techniques render lower-resolution versions of in-game assets when they are far from the camera and progressively replace them with higher-resolution versions as the camera moves closer. 
While there are approaches that propose LOD for 3DGS in large-scale scenes \cite{liu2024citygs,kerbl2024h3dgs,seo2024flod,ren2024octree},
they primarily focus on improving rendering speed, without limiting the number of Gaussians loaded in GPU memory,  making rendering on smaller devices a challenge. Such methods often require to recompute the subset of Gaussians that need to be rendered at every new frame, adding overhead to the rendering. 
More importantly,
this requires all Gaussians from all different LODs (even more than in 3DGS \cite{kerbl20233dgs}) to be in GPU memory at all time.
Finally, existing LOD approaches~\cite{ren2024octree,seo2024flod} require careful parameter tuning for each scene to achieve good quality and performance.

On the other hand, our method is designed to both improve the
rendering speed for large-scale scenes and to limit the number of Gaussians
needed in memory to enable applications on embedded devices. Similarly to existing LOD based methods, we represent the scene as multiple sets of Gaussians with varying level of detail. However, we propose to define regions in space around a cluster center. 
Each region  activates a fixed set of Gaussians from the precomputed LODs to avoid overhead computation between different frames.   
Our contributions can be summarized as follows:
{\setlength{\leftmargini}{2em}
\begin{compactitem}
\item We propose a novel LOD representation for 3DGS which, unlike previous methods \cite{ren2024octree,seo2024flod,liu2024citygs}, does not recompute the list of used Gaussians at each frame and thus can be accelerated and compacted to allow rendering of large-scale scenes even on mobile devices.
\item We further design a strategy to automatically select optimal hyperparameters for splitting LODs, whereas most other methods require hyperparameters to be tuned manually for each 3D scene.
\item To accelerate rendering further, we split the scene into chunks and pre-compute sets of active Gaussians per chunk.
\item Finally, we introduce a novel opacity interpolation scheme to produce visually pleasing rendering and remove any artifacts when transitioning between chunks.
\end{compactitem}}
We show that our method outperforms state-of-the-art (SOTA) approaches in terms of 
quality and rendering speed whilst reducing the number of Gaussians in memory.

\begin{figure}[tbp]
\includegraphics[width=\textwidth]{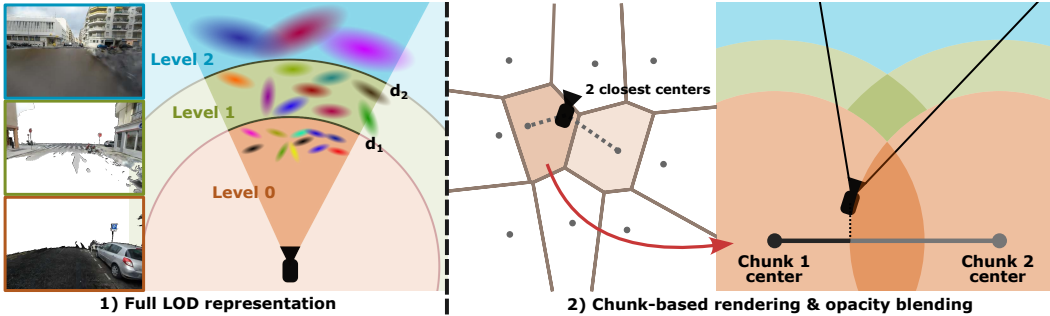}
\caption{\textbf{Method overview.}
\textbf{Left:} The scene is represented with multiple LODs; `active Gaussians' are selected during training based on camera distance.
\textbf{Right:} We cluster cameras into chunks, pre-compute `active Gaussians' per chunk, and render the two nearest chunks with `opacity blending'.
\label{fig:method-overview}}
\end{figure}

\section{Related Work}
\textbf{Novel view synthesis} has received great attention recently, mainly thanks to the advent of neural radiance fields (NeRFs)~\cite{mildenhall2021nerf} and later 3D Gaussian Splatting (3DGS)~\cite{kerbl20233dgs}, which offered an on par alternative with substantially faster rendering.
Subsequently, 3DGS was further extended in many ways to handle antialiasing~\cite{yu2023mip3dgs,lu2023scaffoldgs,shen2025lodgs}, to offer better geometry~\cite{yu2024gof,huang20242dgs}, or to accommodate different camera models~\cite{liao2024fisheye,wu20243dgut}.
There has been a lot of effort on making 3D Gaussian Splatting representation smaller by compressing attributes of Gaussians \cite{lu2023scaffoldgs,fan2024lightgaussian,girish2024eagles,niedermayr2024compressed}.
Similarly to our approach, this also reduces the required memory 
and makes rendering faster. Other approaches modify
the rendering procedure to make rendering more efficient \cite{mallick2024taming3dgs,lin2025metasapiens,kotovenko2025edgs,feng2024flashgs,ye2025gsplat,zoomers2025progs}. 
Finally, a lot of interest has been directed towards optimizing scene representation from images captured under different time-of-day, seasons, or exposure levels \cite{kulhanek2024wildgaussians,niemeyer2024radsplat,lin2024vastgaussian,zhang2024gsw}.
We consider these approaches orthogonal to ours 
as they can also be applied on top of our method.
Similarly, some works \cite{mallick2024taming3dgs,liu2025efficientgs,niemeyer2024radsplat} focused on removing the number of Gaussians 
by proposing alternative densification and pruning strategies \cite{mallick2024taming3dgs,liu2025efficientgs}, or by pruning
Gaussians with little/no contribution to the rendering process \cite{girish2024eagles,fan2024lightgaussian, niemeyer2024radsplat} - similar to our approach.
While these methods make rendering efficient for small-scale scenes, very large scenes
are still not tractable to render in real-time without sacrificing quality.

\PAR{Reconstructing large 3D scenes}
When dealing with extremely large scenes that can span several city blocks, another line of works further proposed to split the scene into sub-regions, and reconstructing each
one separatedly. First, as for NeRFs, \cite{tancik2022blocknerf,turki2022meganerf,zhang2024supernerf} used per-region MLPs and special rendering 
procedures to better handle far-away regions.
Later, similar techniques were also applied in the context of 3DGS~\cite{liu2024citygs,gao2025citygsx,kerbl2024h3dgs,liu2024citygs2}.
These methods can be used to train from large image collections efficiently. Our LOD representation can be built from an existing model and can thus be applied on top of these approaches. 
However, these approaches use a large number of Gaussians in GPU memory and cannot be rendered on low-end devices -- a problem addressed by our method.

\PAR{Level-of-detail 3DGS}
The main contribution of our work is a novel level-of-detail (LOD) representation. Nevertheless, we are not the first to propose incorporating LODs into 3DGS~\cite{seo2024flod,kerbl2024h3dgs,milef2025clod,ren2024octree,liu2024citygs,wang2024pygs}. 
While H3DGS~\cite{kerbl2024h3dgs} and CLOG~\cite{milef2025clod} build continuous multi-level representations, FLoD~\cite{seo2024flod} and OctreeGS~\cite{ren2024octree} use discrete levels, each being a set of Gaussians (similar to our approach). 
The authors start from a coarse set of Gaussians, which are then progressively densified to obtain the finer levels.
Unlike them, our method builds a LOD structure on top of the standard reconstruction and can be applied to various existing methods while we observed that coarse-to-fine strategy tends to fail on large-scale scenes as the
densification fails when the coarse set is too sparse.
Finally, all existing LOD methods \cite{seo2024flod,kerbl2024h3dgs,ren2024octree,milef2025clod} dynamically select the set of `active Gaussians' (Gaussians used in the rasterization) for each rendered frame. This leads to slower rendering and requires all Gaussians to be loaded in GPU memory. 
Unlike them, our method splits the 3D scene into sub-regions and for each, it pre-select sets of `active Gaussians', increasing rendering speed and lowering memory usage.

\section{Method}\label{sec:method}
Our method aims to make large-scale 3D Gaussian Splatting reconstruction
render fast even on mobile devices. To this end, we introduce a novel level-of-detail (LOD) representation and chunk-based caching scheme (see \Cref{fig:method-overview}), leading to fast rendering whilst reducing the memory footprint.

\PAR{Preliminaries: 3D Gaussian Splatting}
3D Gaussian Splatting \cite{kerbl20233dgs} represents a 3D scene as a set of Gaussian primitives, each having a mean (position) $\mu$,
covariance matrix $\Sigma$, opacity $o$, and view-dependent color. The Gaussians are splatted into 2D screen space (obtaining 2D means $\mu'$ and covariances $\Sigma'$~\cite{kerbl20233dgs,zwicker2001surfacesplatting}), sorted in z-axis,
and composited using alpha blending. The blending weight $\alpha$ for pixel $\mathbf{p}$ of 2D Gaussian with mean $\mu'$, covariance $\Sigma'$, and opacity $o$ is then given by the Gaussian function:
\begin{equation}\label{eq:3dgs}
\alpha = o G(\mathbf{p}) \,, \quad\quad\quad G(\mathbf{p}) = e^{-\frac{1}{2}(\mathbf{p} - \mathbf{\mu'})^T (\Sigma')^{-1} (\mathbf{p} - \mathbf{\mu'})} \,.
\end{equation}
Upon splatting of the 3D Gaussians, the 2D screen is split into tiles of $16 \time 16$ pixels each, with each 2D Gaussian being assigned to all (possibly many)
tiles with which it overlaps. Subsequently, the Gaussians for each tile are then sorted and alpha blended in the rasterization process using \Cref{eq:3dgs}.

\PAR{Preliminaries: Importance pruning}
As part of our LOD representation, we adapt the importance pruning introduced in RadSplat~\cite{niemeyer2024radsplat}.
During 3DGS training, many Gaussians can become less visible when their opacity decreases or when
a Gaussian in front of them becomes opaque. While 3DGS periodically removes Gaussians with low opacity,
it does not remove occluded Gaussians (hidden
behind other scene geometry). In RadSplat \cite{niemeyer2024radsplat}, the authors propose to measure each Gaussian's importance (importance score $\tau_i$)
by taking the maximum over the Gaussian's contribution (rendering weight in alpha blending) to any pixel in all training cameras.
By removing all Gaussians with importance score lower than some threshold, we can effectively
prune Gaussians with very little impact on rendering. This makes rendering faster and
also reduces memory, which is particularly important when working with low-end devices.
We follow \cite{niemeyer2024radsplat} and prune twice during training.

\PAR{Level of Detail (LOD) representation}
We propose to represent 3D scenes as multiple sets of Gaussians that correspond to different levels of detail as visualized in \Cref{fig:method-overview}. As motivated in the introduction, distant regions of the scene are often represented by a large number of Gaussians that contribute little to the final rendering. Yet, these Gaussians still need to be evaluated by the renderer, causing excessive memory usage and computations. 
Individual pixels with a large number of visible Gaussians (Gaussians projected into them) can significantly slow down rendering due to how the pixels are processed in $16 \times 16$ patches. The main goal is, therefore, to reduce the number of pixels with a large number of visible Gaussians, as shown in \Cref{fig:lod-histogram}.

Therefore, we propose to represent faraway regions with less detailed sets of Gaussians and nearby regions with more detailed sets. 
To this end, we define multiple LOD levels: $\mathcal{G}^{(l)}; L > l > 0$, with $\mathcal{G}^{(0)}$ denoting the most detailed set, obtained by optimizing on the original set of images~\cite{kerbl20233dgs,kerbl2024h3dgs}.
We assume each LOD level $\mathcal{G}^{(l)}$ is constructed (as explained in the next section) such that a sufficient rendering quality is achieved when observed from a distance of at least $d_l$.
When rendering the LOD from a given camera pose, we select a subset of Gaussians from each LOD level (see \Cref{fig:method-overview}).
We call this set the `active Gaussians', and for camera center $\mathbf{c}$, it is defined as \begin{equation}\label{eq:active-gaussians}
\tilde{\mathcal{G}}(\mathbf{c}) = \bigcup_{l=0}^{L-1} \left\{ g_i \in \mathcal{G}^{(l)} : d_l \le \, \|\mu^{(l)}_i - \mathbf{c}\|_2 < d_{l+1} \right\} \,,
\end{equation}
where $d_0 = 0$, $d_L = \infty$, and $\mu^{(l)}_i$ is the mean of Gaussian $g_i$ from level $l$.
As shown in \Cref{fig:lod-histogram}, LOD rendering reduces long tail of the per-pixel visible Gaussians distribution accelerating the rendering.

\PAR{Building LOD representation}
We aim to construct each LOD level $\mathcal{G}^{(l)}$ such that a sufficient rendering quality is achieved when observed from a distance of at least $d_l$.
To this end, we draw inspiration from the 3D filter proposed by MipSplatting \cite{yu2023mip3dgs}.
As described there, while the rendered image is a 2D projection of a continuous 3D scene given by a set of Gaussians, the image itself is instead a grid in which each pixel coordinate defines the sampling from the continuous 3D signal. For a sampling interval of 1 in screen space, the pixel interval in 3D world space at depth $d$ is $T = \frac{d}{f}$, with $f$ being the focal length. 
According to the Nyquist's theorem \cite{nyquist1928,shannon1949,yu2023mip3dgs}, %
components of the signal can only be reconstructed if they are sampled at invervals smaller than $2T$.
Therefore, Gaussians smaller than $2T$ will only result in aliasing artifacts \cite{yu2023mip3dgs} and increase both memory usage and rendering time. %
To enforce Gaussians of size larger than $2T$, we follow Mip-Splatting~\cite{yu2023mip3dgs} and convolve each Gaussian with a smoothing 3D filter. The resulting Gaussian function (for a Gaussian with mean $\mathbf{\mu}$ and covariance $\Sigma$ and for depth $d$) is given by
\begin{equation}\label{eq:filter}
\tilde{G}(\mathbf{x}) = \sqrt{\frac{|\Sigma|}{|\Sigma + \frac{s d}{f} \cdot \mathbf{I}|}} e^{-\frac{1}{2}(\mathbf{x} - \mathbf{\mu})^T (\Sigma + \frac{s d}{f} \cdot \mathbf{I})^{-1} (\mathbf{x} - \mathbf{\mu})} \,,
\end{equation}
where $s$ is a hyperparameter.

To build the lower-detail representations $\mathcal{G}^{(l)}$ that are viewed at distances larger than $d_l$, we copy Gaussians from $\mathcal{G}^{(0)}$ and add a smoothing 3D filter (\Cref{eq:filter}) for depth $d_l$. While adding a smoothing 3D filter alone does not directly lead to fewer Gaussians, a significant number of Gaussians will become redundant, reducing their contributions in alpha compositing. Eventually, using the importance score from~\cite{niemeyer2024radsplat}, we iteratively prune unused Gaussians. Note that we always employ a few fine-tuning steps to correct for errors introduced by pruning Gaussians. For the fine-tuning, we use LOD rendering with levels up to the currently optimized LOD level.
More details in the \suppmat

\begin{figure}[t!]
    \begin{minipage}{0.64\textwidth}
        \centering
        \includegraphics[height=88pt]{assets/lod-histogram}
        \captionof{figure}{\textbf{Visible Gaussians histogram.} For each pixel we compute the number of visible Gaussians and show the histogram for base model `Full representation' and LOD rendering `LOD'.
        \label{fig:lod-histogram}
        }
    \end{minipage}\hfill
    \begin{minipage}{0.32\textwidth}
        \centering
        \includegraphics[height=88pt]{assets/lod-cost-function-2d}
        \captionof{figure}{\textbf{LOD threshold cost function.} Visualized for 2 depth thresholds. Darker is lower cost.
        \label{fig:lod-cost-function-2d}
        }
        \vspace{-11pt}
    \end{minipage}
\end{figure}

\PAR{Selecting depth thresholds}
An important question remains: how should the depth thresholds $d_l$ be selected to maximize the rendering performance.
The renderer operates on pixels grouped in $16 \times 16$ tiles and the rendering performance is heavily influenced by the number of Gaussians processed inside the same $16 \times 16$-pixels tile. This is because all threads within a tile process the union of all visible Gaussians and must wait for the slowest thread to complete rasterizing.
While it is difficult to theoretically analyze the impact of thresholds on the number of processed Gaussians,
we can estimate this number (cost) for different thresholds - by rendering a subset of training views - and choose the one that minimizes the average number of Gaussians per tile.
In \Cref{fig:lod-cost-function-2d}, we show the cost distribution for two LOD levels at depths $d_1$ and $d_2$. Notice how the value of the minimum is similar for set $\{(x, ax + b) : x \in \mathbb{R}\}$. This enables us to leverage a simple greedy strategy - starting at $d_1$, iteratively adding more thresholds.
This reduces the complexity of the search problem to a linear one.

\PAR{Reducing memory with chunk-based rendering}
While LOD rendering (described so far) makes rasterization fast by reducing the number of visible Gaussians, all Gaussians still need to be loaded in GPU memory. Moreover, the `active Gaussians' need to be continuously recomputed, leading to overhead computations. This can be prohibitive for small devices where the amount of available memory is limited. To address this, we propose to split the scene into different regions called `chunks' and store a fixed set of `active Gaussians' per chunk. 
It is important to note that the set of `active Gaussians' associated with a chunk is not merely a subset of all Gaussians located within its spatial boundaries (see \Cref{fig:method-overview}. Instead, it represents the entire scene, with regions closer to the chunk center modeled at higher levels of detail.
When rendering an image, the rasterizer simply uses the `active Gaussians' pre-computed for the closest chunk.

To define the chunks, we split the scene by performing a K-means clustering on the training camera positions (more details in \suppmat).
For each chunk with center $\mathbf{c}$, the `active Gaussians' are defined by \Cref{eq:active-gaussians} computed at the chunk center (rather than the camera position), with depths $d_l$ offset by the chunk radius (distance to next closest chunk center) to ensure sufficient resolution for all camera positions inside the chunk.

\PAR{Visibility filtering}
Given our LOD chunks, we further accelerate rendering by filtering the set of Gaussians for each LOD chunk.
Following RadSplat~\cite{niemeyer2024radsplat} we additionally perform importance pruning
for each LOD chunk, \ie, we compute per-Gaussian importance scores and remove all Gaussians with importance score lower than 
a fixed threshold.
However, to further increase robustness, we add additional views by adding random perturbations to existing training views within the LOD chunk.
To this end, we employ the original camera positions, but sample random orientations.
Note, that if orientations were included alongside positions in defining LOD chunks, visibility filtering could achieve an additional reduction in the number of loaded Gaussians. Nonetheless, this gain would not be significant and would reduce the method’s ability to handle rapid camera rotations and wide fields of view, both of which are common in practical applications (e.g. AR/VR).

\PAR{Opacity blending for smooth cross-chunk transitions} Having constructed these chunks, a naive approach would be to simply render the `active Gaussians' for the chunk center. While this speeds up rendering and reduces memory, it also introduces sharp changes when rendering a dynamic video of a camera trajectory (see \Cref{fig:ablation}). These artifacts are inherently caused by an abrupt change in the `active Gaussians' while moving, without employing any form of smoothing filter during the transition. To resolve the issue, we propose a smoothing strategy leveraging the two closest chunks (see \Cref{fig:method-overview}). When rendering an image, we first seek the two closest chunk centers. We take their respective sets of active Gaussians and combine them. Afterwards, we modulate the opacity of Gaussians, which are not in the intersection of the two sets of active Gaussians, using
\begin{equation}
\hat{\alpha}_i = \alpha_i t\,, \quad\quad\quad t = \min(1, \max(0, \bar{t}))\,, \quad\quad\quad \bar{t} = \frac{(\mathbf{c} - \mathbf{m}_o)^T(\mathbf{m}_f - \mathbf{m}_o)}{\|\mathbf{m}_o - \mathbf{m}_f\|_2^2}\,,
\end{equation}
where $\mathbf{c}$ is the current camera position, $\mathbf{m}_f$ and  $\mathbf{m}_o$ are the two closest chunk centers -- $\mathbf{m}_f$ being the chunk center to which Gaussian $i$ belongs, and $\mathbf{m}_o$ being the other chunk (not containing Gaussian i). Note that $\bar{t}$ is the normalized length of the projection of $(\mathbf{c} - \mathbf{m_o})$ onto the line connecting the two chunk centers. We use the length of projection instead of the Euclidean distance to achieve a smooth transition even when the camera doesn't pass through the chunk center. Given the union of the two sets of active Gaussians with modified opacity, we proceed with the standard rasterization step. 
However, note that only the union of the two chunks needs to be loaded in memory and at each rendering pass, we only need to update the opacity of the 
symmetric difference of the two active Gaussian sets.

In practice, reloading of LOD splits can be done in a background process that does not affect the renderer's runtime. We begin by loading the active Gaussians’ properties (e.g., positions, colors) into GPU memory. As the camera moves, we only need to modulate the opacity of the two loaded chunks until the camera passes the center of the chunk. Then, we remove the Gaussians from the previous chunk, keep the ones from the closest one, and load Gaussians from the second closest chunk. The opacity blending at this point assigns weights close to 1 to all Gaussians from the closest chunk so there are no artifacts during loading the next chunk -- even if loading of the next chunk is delayed.

\section{Experiments}
To validate our method, we conduct experiments on large-scale indoor and outdoor datasets, comparing our method to other SOTA approaches. We further analyze individual components of our method and evaluate rendering speed on various mobile devices.
We report standard PSNR, SSIM, and LPIPS (VGG) metrics, but also FPS and the number of Gaussians loaded in GPU memory (\#G).
The number of loaded Gaussians serves as a proxy for memory usage, offering a fairer comparison across 3DGS implementations (as opposed to peak-memory usage) as it does not depend on the actual implementations but rather on the algorithm itself.
For more details on implementation and the evaluation protocol, we kindly refer the reader to the \suppmat
All baselines and our method use a single NVIDIA A100 SXM4 40GB GPU for training and evaluation. For mobile experiments, we used two iPhones (13 Mini and 15 Pro), as well as two lower-end laptops without a powerful GPU (MacBook Air M3 and HP Chromebook).

\begin{table}[t]
\addtolength{\tabcolsep}{-0.35em}
\centering
\newcommand{\pf}[1]{\cellcolor{red!30}#1}
\newcommand{\ps}[1]{\cellcolor{orange!30}#1}
\newcommand{\pt}[1]{\cellcolor{yellow!30}#1}
\begin{tabular}{l@{\hskip 6pt}cccrr@{\hskip 12pt}cccrr}
\toprule     
              & \multicolumn{5}{c}{\textbf{SmallCity}} & \multicolumn{5}{c}{\textbf{Campus}} \\
              & PSNR & SSIM & LPIPS & \multicolumn{1}{c}{\#G} & \multicolumn{1}{c}{FPS} & PSNR & SSIM & LPIPS& \multicolumn{1}{c}{\#G} & \multicolumn{1}{c}{FPS}  \\
\midrule
Zip-NeRF~\cite{barron2023zipnerf} & \textit{26.30} & \textit{0.785} & \textit{0.368} & \multicolumn{1}{c}{--} & \textit{0.09} & \textit{22.04} & \textit{0.781} & \textit{0.416} & \multicolumn{1}{c}{--} & \textit{0.20} \\ 
3DGS~\cite{kerbl20233dgs}           &    25.42 &    0.776 &    0.394 & 1375K    &     85.25 &    24.14 &    0.785 &    0.430 &\pt{1142K}&     47.38 \\
Mip-Splatting~\cite{yu2023mip3dgs}  &    25.36 &    0.775 &    0.394 & 1445K    &     75.42 &    23.96 &    0.784 &    0.430 &    1188K &     66.79 \\
Scaffold-GS~\cite{lu2023scaffoldgs} &    23.31 &    0.753 &    0.362 & 1347K    &     71.38 &    20.43 &    0.754 &    0.436 &    1335K &     46.84 \\
H3DGS~\cite{kerbl2024h3dgs}         &\ps{26.42}&\ps{0.807}&\pt{0.331}& 7093K    &     38.07 &    24.60 &\pt{0.798}&\ps{0.396}&    6186K &     34.32 \\
FLOD~\cite{seo2024flod}             &    24.82 &    0.758 &    0.429 &\pf{497K} &\ps{208.41}&    24.10 &    0.777 &    0.453 &\pf{595K} &\pt{120.61}\\
OctreeGS~\cite{ren2024octree}       &\pt{25.98}&\ps{0.807}&\ps{0.326}&\pt{1008K}&\pt{120.27}&\pf{25.22}&\ps{0.800}&\pt{0.408}&\ps{642K} &    119.21 \\
CityGS~\cite{liu2024citygs}         &    25.29 &    0.772 &    0.401 &    2615K &    114.07 &\ps{24.82}&    0.794 &    0.419 &    1881K &\ps{121.67}\\
\Ours                               &\pf{26.57}&\pf{0.815}&\pf{0.325}&\ps{877K} &\pf{257.46}&\pt{24.75}&\pf{0.803}&\pf{0.394}&    1464K &\pf{218.96}\\ %
\bottomrule
\end{tabular}
\vspace{6pt}
\caption{\textbf{Hierarchical 3DGS comparison.}
We compare baselines on SmallCity and Campus scenes. \Ours{}
 achieves fastest rendering while outperforming others in terms of rendering quality.
The \colorbox{red!40}{first}, \colorbox{orange!50}{second}, and \colorbox{yellow!50}{third} values are highlighted.
\label{tab:h3dgs}}
\end{table}

\begin{figure}[t]
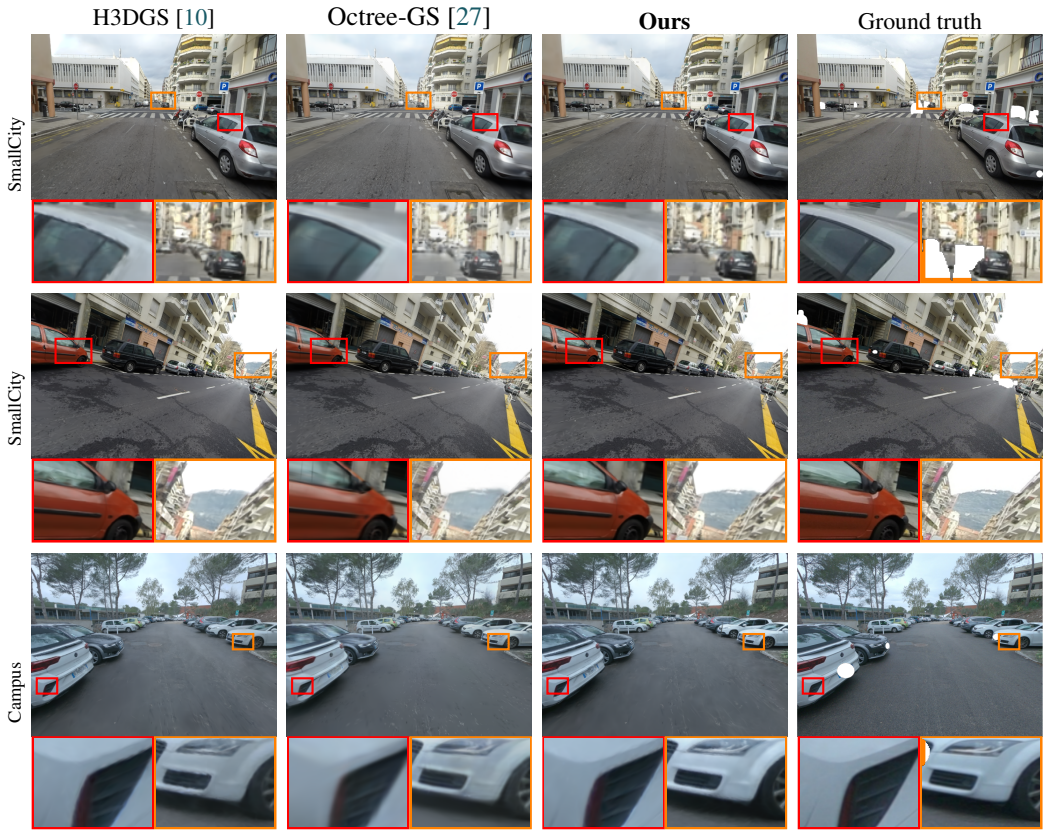

\newcommand\imageHighlightBox[7]{
    \pgfmathsetmacro{\tmpimageWidth}{#5}
    \pgfmathsetmacro{\tmpimageHeight}{#6}
    \pgfmathsetmacro{\tmpcropx}{#1 / \tmpimageWidth}
    \pgfmathsetmacro{\tmpcropy}{#2 / \tmpimageHeight}
    \pgfmathsetmacro{\tmpcropWidth}{#3 / \tmpimageWidth}
    \pgfmathsetmacro{\tmpcropHeight}{#4 / \tmpimageHeight}
    \pgfmathsetmacro{\tmpxtwo}{\tmpcropx + \tmpcropWidth}
    \pgfmathsetmacro{\tmpytwo}{\tmpcropy + \tmpcropHeight}
    \draw[#7, thick] (\tmpcropx,\tmpcropy) rectangle (\tmpxtwo,\tmpytwo);
}

\pgfmathsetmacro{\imageWidth}{1024}
\pgfmathsetmacro{\imageHeight}{690}
\pgfmathsetmacro{\scaleFactorX}{0.24 * \textwidth / \imageWidth}
\pgfmathsetmacro{\scaleFactorY}{0.24 * \textwidth / \imageHeight}
\pgfmathsetmacro{\cropWidth}{100 / \imageWidth}
\pgfmathsetmacro{\cropx}{780 / \imageWidth}
\pgfmathsetmacro{\cropy}{290 / \imageHeight}
\pgfmathsetmacro{\scropx}{500 / \imageWidth}
\pgfmathsetmacro{\scropy}{380 / \imageHeight}

\begin{tikzpicture}
    \newcommand\cscale{0.233}
    \node[] (image) at (0,0) {};
    \newcommand\expandImage{
        \begin{scope}[shift={(image.south west)}]
        \begin{scope}[shift={(image.south west)},x={(image.south east)},y={(image.north west)}]
            \pgfmathsetmacro{\xtwo}{\cropx + \cropWidth}
            \pgfmathsetmacro{\ytwo}{\cropy + \cropWidth}
            \draw[red, thick] (\cropx,\cropy) rectangle (\xtwo,\ytwo);
            \pgfmathsetmacro{\xtwo}{\scropx + \cropWidth}
            \pgfmathsetmacro{\ytwo}{\scropy + \cropWidth}
            \draw[orange, thick] (\scropx,\scropy) rectangle (\xtwo,\ytwo);
        \end{scope}
        \pgfmathsetmacro{\abscropx}{\cropx * \imageWidth}
        \pgfmathsetmacro{\abscropy}{\cropy * \imageHeight}
        \pgfmathsetmacro{\cropt}{(1 - (\cropx + \cropWidth)) * \imageWidth}%
        \pgfmathsetmacro{\cropl}{(1 - (\cropy + \cropWidth)) * \imageHeight}
    
        \node[anchor=north west,inner sep=0,draw=red,line width=2pt] (bottom_image) at (image.south west) {
         \includegraphics[width=0.1115\textwidth,trim={{\abscropx} {\abscropy} {\cropt} {\cropl}},clip]{\imagePath}};
         
        \pgfmathsetmacro{\abscropx}{\scropx * \imageWidth}
        \pgfmathsetmacro{\abscropy}{\scropy * \imageHeight}
        \pgfmathsetmacro{\cropt}{(1 - (\scropx + \cropWidth)) * \imageWidth}%
        \pgfmathsetmacro{\cropl}{(1 - (\scropy + \cropWidth)) * \imageHeight}
        \node[anchor=north east,inner sep=0,draw=orange,line width=2pt] at (image.south east) {
         \includegraphics[width=0.1115\textwidth,trim={{\abscropx} {\abscropy} {\cropt} {\cropl}},clip]{\imagePath}
        };
        \end{scope}
    }

    \def\imagePath{assets/smallcity/h3dgs-pass2_1420}
    \node[anchor=north west,inner sep=0,alias=image,line width=0pt] (image1) at (0,0) {\includegraphics[width=\cscale\textwidth]{\imagePath}};
    \node[anchor=south,yshift=-15pt,xshift=1pt,rotate=90] at (image1.west) 
    {{\scriptsize{} SmallCity}};
    \node[anchor=south, inner sep=2pt] at (image.north) {\small H3DGS~\cite{kerbl2024h3dgs}};
    \expandImage
    \node[anchor=south west] (row-marker) at (bottom_image.south west) {};

    \def\imagePath{assets/smallcity/octreegs-pass2_1420}
    \node[anchor=south west,inner sep=0,alias=image,line width=0pt,xshift=0.01\textwidth] (image2) at (image.south east) {\includegraphics[width=\cscale\textwidth]{\imagePath}};
    \node[anchor=south, inner sep=2pt] at (image.north) {Octree-GS~\cite{ren2024octree}};
    \expandImage
    
    \def\imagePath{assets/smallcity/ours-pass2_1420}
    \node[anchor=south west,inner sep=0,alias=image,line width=0pt,xshift=0.01\textwidth] (image3) at (image.south east) {\includegraphics[width=\cscale\textwidth]{\imagePath}};
    \node[anchor=south, inner sep=2pt] at (image.north) {\small \textbf{Ours}};
    \expandImage

    \def\imagePath{assets/smallcity/gt-pass2_1420}
    \node[anchor=south west,inner sep=0,alias=image,line width=0pt,xshift=0.01\textwidth] (image4) at (image.south east) {\includegraphics[width=\cscale\textwidth]{\imagePath}};
    \node[anchor=south, inner sep=2pt] at (image.north) {\small Ground truth};
    \expandImage

\pgfmathsetmacro{\imageWidth}{1024}
\pgfmathsetmacro{\imageHeight}{690}
\pgfmathsetmacro{\scaleFactorX}{0.24 * \textwidth / \imageWidth}
\pgfmathsetmacro{\scaleFactorY}{0.24 * \textwidth / \imageHeight}
\pgfmathsetmacro{\cropWidth}{150 / \imageWidth}
\pgfmathsetmacro{\cropx}{100 / \imageWidth}
\pgfmathsetmacro{\cropy}{400 / \imageHeight}
\pgfmathsetmacro{\scropx}{850 / \imageWidth}
\pgfmathsetmacro{\scropy}{340 / \imageHeight}

    \def\imagePath{assets/smallcity/h3dgs-pass3_0271}
    \node[anchor=north west,inner sep=0,alias=image,line width=0pt,yshift=-0.01\textwidth] (image1) at (row-marker.south west) {\includegraphics[width=\cscale\textwidth]{\imagePath}};
    \expandImage
    \node[anchor=south west] (row-marker) at (bottom_image.south west) {};
    \node[anchor=south,yshift=-13pt,xshift=1pt,rotate=90] at (image1.west) 
    {{\scriptsize{} SmallCity}};
    
    \def\imagePath{assets/smallcity/octreegs-pass3_0271}
    \node[anchor=south west,inner sep=0,alias=image,line width=0pt,xshift=0.01\textwidth] (image2) at (image.south east) {\includegraphics[width=\cscale\textwidth]{\imagePath}};
    \expandImage
    
    \def\imagePath{assets/smallcity/ours-pass3_0271}
    \node[anchor=south west,inner sep=0,alias=image,line width=0pt,xshift=0.01\textwidth] (image3) at (image.south east) {\includegraphics[width=\cscale\textwidth]{\imagePath}};
    \expandImage

    \def\imagePath{assets/smallcity/gt-pass3_0271}
    \node[anchor=south west,inner sep=0,alias=image,line width=0pt,xshift=0.01\textwidth] (image4) at (image.south east) {\includegraphics[width=\cscale\textwidth]{\imagePath}};
    \expandImage

    \pgfmathsetmacro{\imageWidth}{1436}
    \pgfmathsetmacro{\imageHeight}{1075}
    \pgfmathsetmacro{\scaleFactorX}{0.24 * \textwidth / \imageWidth}
    \pgfmathsetmacro{\scaleFactorY}{0.24 * \textwidth / \imageHeight}
    \pgfmathsetmacro{\cropWidth}{120 / \imageWidth}
    \pgfmathsetmacro{\cropx}{30 / \imageWidth}
    \pgfmathsetmacro{\cropy}{250 / \imageHeight}
    \pgfmathsetmacro{\scropx}{1180 / \imageWidth}
    \pgfmathsetmacro{\scropy}{510 / \imageHeight}

    \def\imagePath{assets/campus/h3dgs-back_G0042596}
    \node[anchor=north west,inner sep=0,alias=image,line width=0pt,yshift=-0.01\textwidth] (image1) at (row-marker.south west) {\includegraphics[width=\cscale\textwidth]{\imagePath}};
    \expandImage
    \node[anchor=south west] (row-marker) at (bottom_image.south west) {};
    \node[anchor=south,yshift=-19pt,xshift=1pt,rotate=90] at (image1.west) 
    {{\scriptsize{} Campus}};
    
    \def\imagePath{assets/campus/octreegs-back_G0042596}
    \node[anchor=south west,inner sep=0,alias=image,line width=0pt,xshift=0.01\textwidth] (image2) at (image.south east) {\includegraphics[width=\cscale\textwidth]{\imagePath}};
    \expandImage
    
    \def\imagePath{assets/campus/ours-back_G0042596}
    \node[anchor=south west,inner sep=0,alias=image,line width=0pt,xshift=0.01\textwidth] (image3) at (image.south east) {\includegraphics[width=\cscale\textwidth]{\imagePath}};
    \expandImage

    \def\imagePath{assets/campus/gt-back_G0042596}
    \node[anchor=south west,inner sep=0,alias=image,line width=0pt,xshift=0.01\textwidth] (image4) at (image.south east) {\includegraphics[width=\cscale\textwidth]{\imagePath}};
    \expandImage

\end{tikzpicture}
\caption{\textbf{Hierarchical 3DGS qualitative results.} We compare H3DGS~\cite{kerbl2024h3dgs} and Octree-GS~\cite{ren2024octree} on SmallCity and Campus scenes. We highlight \textcolor{red}{close-up region (left)} and \textcolor{orange}{far-away region (right)}.
\label{fig:h3dgs}}
\end{figure}

\PAR{Datasets \& baselines}
We use two larger-scale datasets to validate our approach, \ie, two outdoor scenes from Hierarchical 3DGS dataset~\cite{kerbl2024h3dgs} and three indoor scenes from Zip-NeRF dataset \cite{barron2023zipnerf}. Each scene consists of around 1000-2000 images.
We compare our approach against the following baselines:
Current SOTA on the Zip-NeRF dataset - \textbf{Zip-NeRF}~\cite{barron2023zipnerf} - slow to train and without real-time rendering.
Traditional 3DGS baselines - \textbf{3DGS}~\cite{kerbl20233dgs}, \textbf{Mip-Splatting}~\cite{yu2023mip3dgs}, and \textbf{Scaffold-GS}~\cite{lu2023scaffoldgs}.
LOD-based methods - \textbf{H3DGS}~\cite{kerbl2024h3dgs} (SOTA on the Hierarchical 3DGS dataset), \textbf{FLOD}~\cite{seo2024flod}, \textbf{Octree-GS}~\cite{ren2024octree}, and \textbf{CityGS}~\cite{liu2024citygs}. Note that H3DGS, Octree-GS, FLOD were trained for 45K, 40K, and 100K iterations, while \ours{} was trained for 36K iterations. For FLOD we use 3-4-5 LOD rendering and for H3DGS we employ their default $\tau = 6$. More details can be found in the \suppmat

\PAR{Evaluation on Hierarchical 3DGS dataset}
As shown in~\Cref{tab:h3dgs}, Zip-NeRF struggles on these large scenes, despite requiring much longer training (200K iterations vs. 36K for \ours). 
This likely stems from its use of a contiguous scene representation with limited resolution, unlike 3DGS, which sparsely encodes only occupied space.
Among non-LOD baselines, 3DGS~\cite{kerbl20233dgs} and Mip-Splatting~\cite{yu2023mip3dgs} offer competitive quality and rendering speed, whereas Scaffold-GS~\cite{lu2023scaffoldgs} lags behind, likely due to suboptimal densification in large-scale scenes.
H3DGS~\cite{kerbl2024h3dgs} attains high visual fidelity, but incurs significant overhead from per-frame LOD splitting and the large number of Gaussians, making it slower than other methods.
FLOD fails to generate sufficient Gaussians due to their coarse-to-fine training scheme, discarding fine details early and struggling to recover them later, even though it uses three times more training iterations than related methods. 
Compared to Octree-GS~\cite{ren2024octree} and CityGS~\cite{liu2024citygs}, \ours{} achieves higher overall quality while being twice as fast. Interestingly, both Octree-GS and CityGS excel on Campus in PSNR but underperform in SSIM and LPIPS. We attribute this to our method being more robust to exposure variation
but less effective in preserving perceptual similarity under appearance changes.
In summary, our method consistently achieves the best results across all scenes whilst maintaining superior rendering efficiency.
In qualitative comparisons (\Cref{fig:h3dgs}), we highlight a \textcolor{red}{close-up} and \textcolor{orange}{distant} regions to assess both local details and long-range fidelity.
Overall, our method delivers sharper reconstructions in both close and distant regions. In contrast, Octree-GS appears blurrier nearby and suffers from color shifts at a distance, while H3DGS often exhibits jagged geometry, potentially due to its reliance on depth supervision.

\begin{figure}[t!]
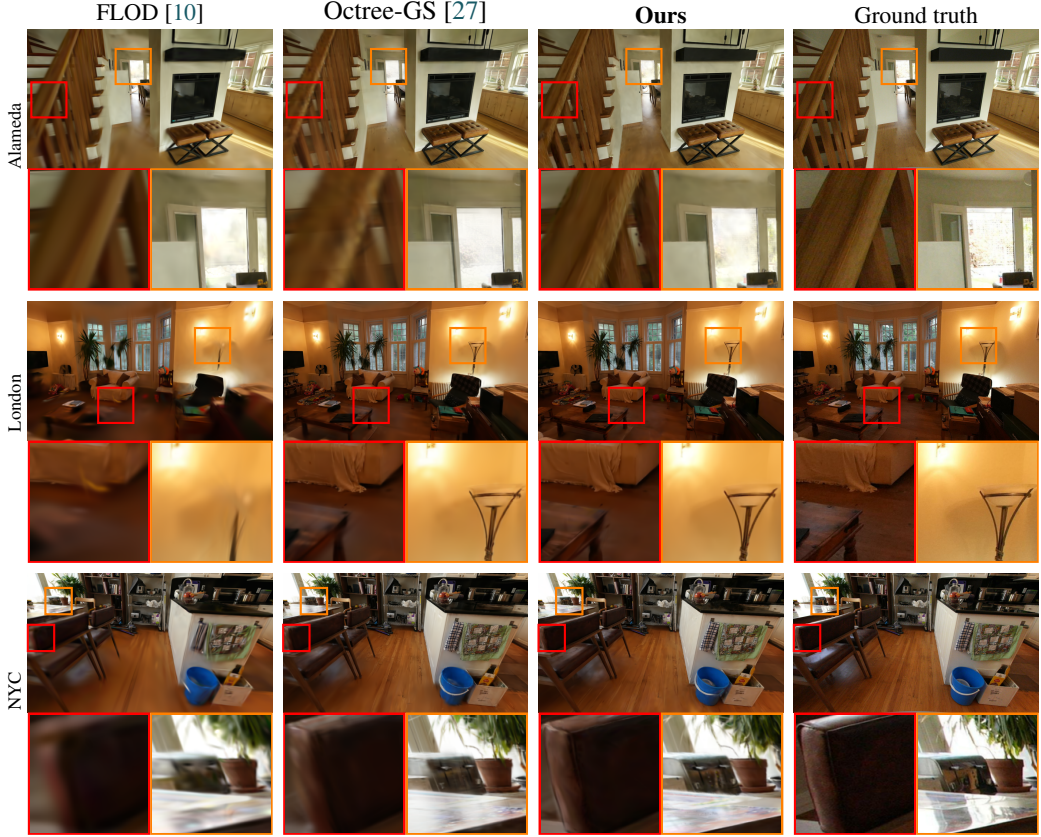

\begin{tikzpicture}
    \newcommand\cscale{0.233}
    \node[] (image) at (0,0) {};
    \newcommand\expandImage{
        \begin{scope}[shift={(image.south west)}]
        \begin{scope}[shift={(image.south west)},x={(image.south east)},y={(image.north west)}]
            \pgfmathsetmacro{\xtwo}{\cropx + \cropWidth}
            \pgfmathsetmacro{\ytwo}{\cropy + \cropHeight}
            \draw[red, thick] (\cropx,\cropy) rectangle (\xtwo,\ytwo);
            \pgfmathsetmacro{\xtwo}{\scropx + \cropWidth}
            \pgfmathsetmacro{\ytwo}{\scropy + \cropHeight}
            \draw[orange, thick] (\scropx,\scropy) rectangle (\xtwo,\ytwo);
        \end{scope}
        \pgfmathsetmacro{\abscropx}{\cropx * \imageWidth}
        \pgfmathsetmacro{\abscropy}{\cropy * \imageHeight}
        \pgfmathsetmacro{\cropt}{(1 - (\cropx + \cropWidth)) * \imageWidth}%
        \pgfmathsetmacro{\cropl}{(1 - (\cropy + \cropHeight)) * \imageHeight}
    
        \node[anchor=north west,inner sep=0,draw=red,line width=2pt] (bottom_image) at (image.south west) {
         \includegraphics[width=0.1115\textwidth,trim={{\abscropx} {\abscropy} {\cropt} {\cropl}},clip]{\imagePath}};
         
        \pgfmathsetmacro{\abscropx}{\scropx * \imageWidth}
        \pgfmathsetmacro{\abscropy}{\scropy * \imageHeight}
        \pgfmathsetmacro{\cropt}{(1 - (\scropx + \cropWidth)) * \imageWidth}%
        \pgfmathsetmacro{\cropl}{(1 - (\scropy + \cropHeight)) * \imageHeight}
        \node[anchor=north east,inner sep=0,draw=orange,line width=2pt] at (image.south east) {
         \includegraphics[width=0.1115\textwidth,trim={{\abscropx} {\abscropy} {\cropt} {\cropl}},clip]{\imagePath}
        };
        \end{scope}
    }

    \pgfmathsetmacro{\imageWidth}{1394}
    \pgfmathsetmacro{\imageHeight}{793}
    \pgfmathsetmacro{\scaleFactorX}{0.24 * \textwidth / \imageWidth}
    \pgfmathsetmacro{\scaleFactorY}{0.24 * \textwidth / \imageHeight}
    \pgfmathsetmacro{\cropWidth}{200 / \imageWidth}
    \pgfmathsetmacro{\cropHeight}{200 / \imageHeight}
    \pgfmathsetmacro{\cropx}{20 / \imageWidth}
    \pgfmathsetmacro{\cropy}{290 / \imageHeight}
    \pgfmathsetmacro{\scropx}{500 / \imageWidth}
    \pgfmathsetmacro{\scropy}{480 / \imageHeight}

    \def\imagePath{assets/alameda/flod-indoor_DSC06796}
    \node[anchor=north west,inner sep=0,alias=image,line width=0pt] (image1) at (0,0) {\includegraphics[width=\cscale\textwidth]{\imagePath}};
    \node[anchor=south,yshift=-15pt,xshift=1pt,rotate=90] at (image1.west) 
    {{\scriptsize{} Alameda}};
    \node[anchor=south, inner sep=2pt] at (image.north) {\small FLOD~\cite{kerbl2024h3dgs}};
    \expandImage
    \node[anchor=south west] (row-marker) at (bottom_image.south west) {};

    \def\imagePath{assets/alameda/octreegs-indoor_DSC06796}
    \node[anchor=south west,inner sep=0,alias=image,line width=0pt,xshift=0.01\textwidth] (image2) at (image.south east) {\includegraphics[width=\cscale\textwidth]{\imagePath}};
    \node[anchor=south, inner sep=2pt] at (image.north) {Octree-GS~\cite{ren2024octree}};
    \expandImage
    
    \def\imagePath{assets/alameda/ours-indoor_DSC06796}
    \node[anchor=south west,inner sep=0,alias=image,line width=0pt,xshift=0.01\textwidth] (image3) at (image.south east) {\includegraphics[width=\cscale\textwidth]{\imagePath}};
    \node[anchor=south, inner sep=2pt] at (image.north) {\small \textbf{Ours}};
    \expandImage

    \def\imagePath{assets/alameda/gt-indoor_DSC06796}
    \node[anchor=south west,inner sep=0,alias=image,line width=0pt,xshift=0.01\textwidth] (image4) at (image.south east) {\includegraphics[width=\cscale\textwidth]{\imagePath}};
    \node[anchor=south, inner sep=2pt] at (image.north) {\small Ground truth};
    \expandImage

\pgfmathsetmacro{\imageWidth}{1392}
\pgfmathsetmacro{\imageHeight}{793}
\pgfmathsetmacro{\scaleFactorX}{0.24 * \textwidth / \imageWidth}
\pgfmathsetmacro{\scaleFactorY}{0.24 * \textwidth / \imageHeight}
\pgfmathsetmacro{\cropWidth}{200 / \imageWidth}
\pgfmathsetmacro{\cropHeight}{200 / \imageHeight}
\pgfmathsetmacro{\cropx}{400 / \imageWidth}
\pgfmathsetmacro{\cropy}{100 / \imageHeight}
\pgfmathsetmacro{\scropx}{950 / \imageWidth}
\pgfmathsetmacro{\scropy}{440 / \imageHeight}

    \def\imagePath{assets/london/flod-indoor_DSC01195}
    \node[anchor=north west,inner sep=0,alias=image,line width=0pt,yshift=-0.01\textwidth] (image1) at (row-marker.south west) {\includegraphics[width=\cscale\textwidth]{\imagePath}};
    \expandImage
    \node[anchor=south west] (row-marker) at (bottom_image.south west) {};
    \node[anchor=south,yshift=-13pt,xshift=1pt,rotate=90] at (image1.west) 
    {{\scriptsize{} London}};
    
    \def\imagePath{assets/london/octreegs-indoor_DSC01195}
    \node[anchor=south west,inner sep=0,alias=image,line width=0pt,xshift=0.01\textwidth] (image2) at (image.south east) {\includegraphics[width=\cscale\textwidth]{\imagePath}};
    \expandImage
    
    \def\imagePath{assets/london/ours-indoor_DSC01195}
    \node[anchor=south west,inner sep=0,alias=image,line width=0pt,xshift=0.01\textwidth] (image3) at (image.south east) {\includegraphics[width=\cscale\textwidth]{\imagePath}};
    \expandImage

    \def\imagePath{assets/london/gt-indoor_DSC01195}
    \node[anchor=south west,inner sep=0,alias=image,line width=0pt,xshift=0.01\textwidth] (image4) at (image.south east) {\includegraphics[width=\cscale\textwidth]{\imagePath}};
    \expandImage

    \pgfmathsetmacro{\imageWidth}{1393}
    \pgfmathsetmacro{\imageHeight}{793}
    \pgfmathsetmacro{\scaleFactorX}{0.24 * \textwidth / \imageWidth}
    \pgfmathsetmacro{\scaleFactorY}{0.24 * \textwidth / \imageHeight}
    \pgfmathsetmacro{\cropWidth}{150 / \imageWidth}
    \pgfmathsetmacro{\cropHeight}{150 / \imageHeight}
    \pgfmathsetmacro{\cropx}{0 / \imageWidth}
    \pgfmathsetmacro{\cropy}{350 / \imageHeight}
    \pgfmathsetmacro{\scropx}{100 / \imageWidth}
    \pgfmathsetmacro{\scropy}{560 / \imageHeight}

    \def\imagePath{assets/nyc/flod-indoor_DSC02966}
    \node[anchor=north west,inner sep=0,alias=image,line width=0pt,yshift=-0.01\textwidth] (image1) at (row-marker.south west) {\includegraphics[width=\cscale\textwidth]{\imagePath}};
    \expandImage
    \node[anchor=south west] (row-marker) at (bottom_image.south west) {};
    \node[anchor=south,yshift=-19pt,xshift=1pt,rotate=90] at (image1.west) 
    {{\scriptsize{} NYC}};
    
    \def\imagePath{assets/nyc/octreegs-indoor_DSC02966}
    \node[anchor=south west,inner sep=0,alias=image,line width=0pt,xshift=0.01\textwidth] (image2) at (image.south east) {\includegraphics[width=\cscale\textwidth]{\imagePath}};
    \expandImage
    
    \def\imagePath{assets/nyc/ours-indoor_DSC02966}
    \node[anchor=south west,inner sep=0,alias=image,line width=0pt,xshift=0.01\textwidth] (image3) at (image.south east) {\includegraphics[width=\cscale\textwidth]{\imagePath}};
    \expandImage

    \def\imagePath{assets/nyc/gt-indoor_DSC02966}
    \node[anchor=south west,inner sep=0,alias=image,line width=0pt,xshift=0.01\textwidth] (image4) at (image.south east) {\includegraphics[width=\cscale\textwidth]{\imagePath}};
    \expandImage
\end{tikzpicture}
\caption{\textbf{Zip-NeRF dataset qualitative comparison.} On scenes: Alameda, London, NYC.  We highlight \textcolor{red}{close-up region (left)} and \textcolor{orange}{far-away region (right)}.
\label{fig:zipnerf}}
\end{figure}
\begin{table}[tbp]
\addtolength{\tabcolsep}{-0.3em}
\centering
\newcommand{\pf}[1]{\cellcolor{red!30}#1}
\newcommand{\ps}[1]{\cellcolor{orange!30}#1}
\newcommand{\pt}[1]{\cellcolor{yellow!30}#1}
\begin{tabular}{l@{\hskip 15pt}ccr@{\hskip 15pt}ccr@{\hskip 15pt}ccr}
\toprule
& \multicolumn{3}{c}{\textbf{Alameda}} & \multicolumn{3}{c}{\textbf{London}} & \multicolumn{3}{c}{\textbf{NYC}} \\
Method & PSNR & SSIM & FPS & PSNR & SSIM & FPS & PSNR & SSIM & FPS \\
\midrule
\textit{Zip-NeRF}~\cite{barron2023zipnerf} & \textit{22.97} & \textit{0.738} & \textit{0.13} & \textit{26.76} & \textit{0.822} & \textit{0.13} & \textit{28.21} & \textit{0.845} & \textit{0.13} \\ \hline
3DGS~\cite{kerbl20233dgs}           &    21.75 &    0.707 &  90.12    &    25.43 &    0.795 &    99.19  &    26.33 &    0.829 & 79.69 \\
Mip-Splatting~\cite{yu2023mip3dgs}  &    21.86 &    0.709 &  95.52    &    25.55 &    0.797 &    95.56  &    26.33 &    0.829 & 85.24 \\
Scaffold-GS~\cite{lu2023scaffoldgs} &    20.93 &    0.714 &  65.61    &    22.52 &    0.745 &    75.71  &    25.97 &    0.817 & 78.74 \\
H3DGS~\cite{kerbl2024h3dgs}         &22.21&\ps{0.739}&  27.82    &\pf{26.34}&\pf{0.823}&    30.49  &\ps{27.28}&\pf{0.849}& 33.11 \\
FLOD~\cite{seo2024flod}             &    21.35 &    0.666 &\pf{276.52}&    24.38 &    0.753 &\ps{195.06}&    25.01 &    0.781 &\ps{260.85}\\
OctreeGS~\cite{ren2024octree}       &\pf{22.94}&\pt{0.734}&119.83&\pt{26.04}&\pt{0.817}& 153.06 &\pt{27.05}&\pt{0.839}&\pt{146.29}\\
CityGS~\cite{liu2024citygs}&\ps{22.43}& 0.729 &\pt{174.07}& 25.87 & 0.809 &\pt{190.10}& 26.55 & \pt{0.839} & 144.91 \\
\Ours                               &\pt{22.41}&\pf{0.741}&\ps{229.99}&\pf{26.34}&\ps{0.818}&\pf{252.58}&\pf{27.40}&\pf{0.849}&\pf{280.22}\\
\bottomrule
\end{tabular}
\vspace{6pt}
\caption{\textbf{Zip-NeRF dataset comparison.} On scenes: Alameda, London, NYC. We report PSNR, SSIM, and FPS.
The \colorbox{red!40}{first}, \colorbox{orange!50}{second}, and \colorbox{yellow!50}{third} values are highlighted.
\label{tab:zipnerf}}
\vspace{-3pt}%
\end{table}

\PAR{Zip-NeRF dataset evaluation}
As seen in~\Cref{tab:zipnerf}, Zip-NeRF achieves the highest accuracy but requires long training (200K iterations) and does not support real-time rendering. 
We still include it for reference. 
Among real-time-capable methods, our approach achieves the best trade-off between quality and speed. CityGS~\cite{liu2024citygs} slower than \ours{} while also having worse quality.
FLOD~\cite{seo2024flod} renders fast but loses fine details due to early pruning in its coarse-to-fine training. H3DGS~\cite{kerbl2024h3dgs} maintains high fidelity but is significantly slower, requiring over 10M Gaussians and per-frame LOD graph cuts. Octree-GS~\cite{ren2024octree} offers good quality and decent speed, but is slower than ours due to per-frame MLP evaluations. Finally, Scaffold-GS~\cite{lu2023scaffoldgs} lags behind other non-LOD baselines in both quality and speed.
As for qualitative evaluation (\Cref{fig:zipnerf}), FLOD tends to produce blurrier results and often misses distant details (e.g., lamp in \textcolor{orange}{2$^\text{nd}$ row}). Octree-GS struggles with close-up sharpness (e.g., railing in \textcolor{red}{1$^\text{st}$ row}) compared to our method. For far regions, both methods perform similarly, though our model better preserves high-frequency highlights (e.g., London scene), while Octree-GS slightly reduces color artifacts under exposure variation due to its MLP-based design.
In summary, our method can deliver sharper close-up reconstructions and comparable or superior fidelity for far regions, while maintaining faster rendering speed.%

In the Hierarchical 3DGS dataset~\cite{kerbl2024h3dgs}, the camera follows a linear trajectory with chunks aligned along it; therefore, only the two nearest chunks are relevant. In Zip-NeRF~\cite{barron2023zipnerf}, however, the chunks fully cover the scene and the camera may traverse diagonally, potentially being between three or more chunks -- raising the question of whether opacity blending of only two chunks is sufficient.
Empirically, no artifacts were observed across Hierarchical 3DGS~\cite{kerbl2024h3dgs}, Zip-NeRF~\cite{barron2023zipnerf}, and Mip-NeRF 360~\cite{barron2022mip360}. We hypothesize that:
\textbf{a)} Close to the training camera distribution, neighboring chunks are well constrained, producing near-identical renderings even without blending.
\textbf{b)} Far from the distribution, viewpoints lie at chunk boundaries where only the two closest chunks dominate. Thus, two-chunk blending is generally sufficient.

\begin{table}[tbp]
\centering
\begin{tabular}{lrrrr}
\toprule
                         & PSNR  & Time  & \#Vis. G & \multicolumn{1}{c}{\#G} \\ %
\midrule
\textit{Full representation}             & 26.62 & 15.17 & 925k     & 2639K     \\
LOD (d=10)                                 & 26.45 & 5.61  & 324K     & 3465M     \\
LOD (d=10,28)                             & 26.50 & 4.75  & 209K     & 3815K     \\
LOD (d=10,28,47)                         & 26.46 & 4.17  & 182K     & 4016K    \\
LOD (d=10,28,47,63)                     & 26.50 & 4.07  & 172K     & 4145K    \\
\midrule
LOD (d=15,70)                             & 26.55 & 5.24  & 267K     & 3377K       \\
\midrule
LOD (d=10,28) + clusters                  & 26.54 & 3.62  & 244K     & 795K     \\
LOD (d=10,28) + clusters + vis. filtering & 26.55 & 3.15  & 185K     & 612K     \\
Opacity blending                         & 26.57 & 3.88  & 268K     & 877K     \\
\bottomrule
\end{tabular}
\vspace{6pt}
\caption{\textbf{Performance analysis.}
We show impact of different number of LOD levels and impact of LOD clusters, visibility filtering, 
and opacity blending on rendering time and quality (PSNR).
For LOD, we vary the number of levels and the depths $d$.
Last three rows use chunk-based rendering.
\label{tab:ablation}}
\end{table}

\PAR{Performance analysis \& design validation}
We conduct an ablation study on the SmallCity scene to assess the impact of key components of our method on rendering quality and speed. Quantitative results are shown in \Cref{tab:ablation} (corresponding to SmallCity in \Cref{tab:h3dgs}), reporting PSNR, render time, number of visible Gaussians, and memory usage. Time for reloading Gaussians at chunk boundaries is excluded, as it is handled asynchronously. Qualitative results are shown in \Cref{fig:ablation}.
We start with the base method (`Full representation’), which includes importance pruning but no LOD.
Adding up to four LOD levels significantly improves rendering speed (up to $3\times$ with just one level) at a minimal cost in PSNR and sharpness due to more aggressive pruning. 
As additional levels yield diminishing returns, we decided to always employ two LODs for all experiments, as a good speed and quality trade-off. 
Thresholds were selected automatically (\Cref{sec:method}, `Selecting depth thresholds'), and we show that when instead setting them manually (15 and 70 m), performance worsens compared to our automatic selection `LOD (d=10,28)'.

The last three rows evaluate our chunk-based rendering. 
Clustering camera positions and adjusting depth thresholds by chunk radius improves quality (`LOD (d=10,28) + clusters’), but increases Gaussian count due to finer LOD resolutions. Visibility filtering (`+ vis. filtering’) further reduces both visible and loaded Gaussians, and thus rendering time.
However, using the active Gaussians from only the chunk centers introduces visual artifacts at chunk boundaries, noticeable as sharpness discontinuities (\Cref{fig:ablation}, second column from right). To circumvent this issue, we introduce opacity blending between adjacent chunks (last row, corresponding to `Ours' in \Cref{tab:h3dgs}), which smoothens transitions. While it slightly increases render time and Gaussian count, it runs faster and more memory-efficient than its non-chunk-based LOD counterpart, being particularly beneficial for memory-constrained mobile devices.

\begin{figure}[tbp]
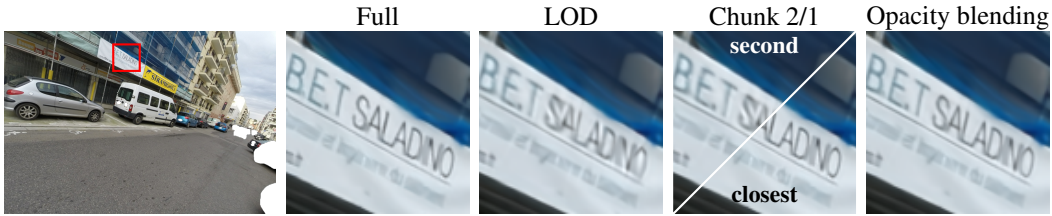

\newcommand\imageHighlightBox[7]{
    \pgfmathsetmacro{\tmpimageWidth}{#5}
    \pgfmathsetmacro{\tmpimageHeight}{#6}
    \pgfmathsetmacro{\tmpcropx}{#1 / \tmpimageWidth}
    \pgfmathsetmacro{\tmpcropy}{#2 / \tmpimageHeight}
    \pgfmathsetmacro{\tmpcropWidth}{#3 / \tmpimageWidth}
    \pgfmathsetmacro{\tmpcropHeight}{#4 / \tmpimageHeight}
    \pgfmathsetmacro{\tmpxtwo}{\tmpcropx + \tmpcropWidth}
    \pgfmathsetmacro{\tmpytwo}{\tmpcropy + \tmpcropHeight}
    \draw[#7, thick] (\tmpcropx,\tmpcropy) rectangle (\tmpxtwo,\tmpytwo);
}

\pgfmathsetmacro{\imageWidth}{1024}
\pgfmathsetmacro{\imageHeight}{690}
\pgfmathsetmacro{\scaleFactorX}{0.24 * \textwidth / \imageWidth}
\pgfmathsetmacro{\scaleFactorY}{0.24 * \textwidth / \imageHeight}
\pgfmathsetmacro{\cropWidth}{100}
\pgfmathsetmacro{\cropHeight}{100}%
\pgfmathsetmacro{\cropx}{410}
\pgfmathsetmacro{\cropy}{\imageHeight-\cropHeight - 50}

\pgfmathsetmacro{\cropt}{(\imageWidth - \cropx - \cropWidth)}
\pgfmathsetmacro{\cropl}{(\imageHeight - \cropy - \cropWidth)}

\begin{tikzpicture}
    \newcommand\cscale{0.174}

    \node[anchor=north west,inner sep=0,alias=image,line width=0pt] (image-gs) at (0,0) {
        \includegraphics[height=\cscale\textwidth]{assets/ablation/ablation-gt-9}};

    \begin{scope}[shift={(image.south west)},x={(image.south east)},y={(image.north west)}]
        \imageHighlightBox{\cropx}{\cropy}{\cropWidth}{\cropHeight}{\imageWidth}{\imageHeight}{red}
    \end{scope}
    
    \node[anchor=south west,inner sep=0,alias=image,line width=0pt,xshift=0.01\textwidth] (image3) at (image.south east) {\includegraphics[height=\cscale\textwidth,trim={{\cropx} {\cropy} {\cropt} {\cropl}},clip]{assets/ablation/ablation-full-9}};
    \node[anchor=south, inner sep=2pt] at (image.north) {Full};

    \node[anchor=south west,inner sep=0,alias=image,line width=0pt,xshift=0.01\textwidth] (image2) at (image.south east) {\includegraphics[height=\cscale\textwidth,trim={{\cropx} {\cropy} {\cropt} {\cropl}},clip]{assets/ablation/ablation-lod-9}};
    \node[anchor=south, inner sep=2pt] at (image.north) {LOD};

    \node[anchor=south west,inner sep=0,alias=image,line width=0pt,xshift=0.01\textwidth,opacity=0] (image4) at (image.south east) {\includegraphics[height=\cscale\textwidth,trim={{\cropx} {\cropy} {\cropt} {\cropl}},clip]{assets/ablation/ablation-c1-9}};
    \node[anchor=south, inner sep=2pt] at (image.north) {Chunk 2/1};
    \begin{scope}
        \clip
          ([yshift=0.2mm]image.north east) --
          ([yshift=0.2mm]image.north west) --
          ([yshift=0.2mm]image.south west) -- cycle;
        \node[anchor=north west,inner sep=0,line width=0pt] at (image.north west) {\includegraphics[height=\cscale\textwidth,trim={{\cropx} {\cropy} {\cropt} {\cropl}},clip]{assets/ablation/ablation-c2-9}};
    \end{scope}
    \begin{scope}
        \clip
          ([yshift=-0.2mm]image.north east) --
          ([yshift=-0.2mm]image.south east) --
          ([yshift=-0.2mm]image.south west) -- cycle;
        \node[anchor=north west,inner sep=0,line width=0pt] at (image.north west) {\includegraphics[height=\cscale\textwidth,trim={{\cropx} {\cropy} {\cropt} {\cropl}},clip]{assets/ablation/ablation-c1-9}};
    \end{scope}
    \node[anchor=north, inner sep=2pt,color=white] at (image.north) {\small{}\textbf{second}};
    \node[anchor=south, inner sep=4pt,color=black] at (image.south) {\small{}\textbf{closest}};
    
    \node[anchor=south west,inner sep=0,alias=image,line width=0pt,xshift=0.01\textwidth] (image4) at (image.south east) {\includegraphics[height=\cscale\textwidth,trim={{\cropx} {\cropy} {\cropt} {\cropl}},clip]{assets/ablation/ablation-blended-9}};
    \node[anchor=south, inner sep=0.3pt] at (image.north) {Opacity blending};
    
\end{tikzpicture}
\caption{\textbf{Ablation study.} We compare the full representation (without LOD), full LOD, rendering from closest and second closest LOD chunks, and opacity blending.
\label{fig:ablation}}
\end{figure}

\newcommand{\fMark}[0]{\textcolor{red}{\xmark}}
\newcommand{\cMark}[0]{\textcolor{OliveGreen}{\cmark}}
\begin{table}[tbp]
\addtolength{\tabcolsep}{-0.35em}
\begin{tabular}{lc@{\hskip 2em}cc@{\hskip 2em}cc@{\hskip 2em}cc@{\hskip 2em}cc}
\toprule
& &
\multicolumn{2}{c}{\hspace{-2em}iPhone 13 Mini\hspace{-2em}} &
\multicolumn{2}{c}{\hspace{-2em}iPhone 15 Pro\hspace{-2em}} &
\multicolumn{2}{c}{\hspace{-2em}MacBook M3\hspace{-2em}} & 
\multicolumn{2}{c}{\hspace{-0.3em}Chromebook} \\
& HQ &
FPS & S. time &
FPS & S. time &
FPS & S. time &
FPS & S. time \\
\midrule
H3DGS                 & \cMark & \fMark & \fMark & \fMark  & \fMark  & 7 & 42 & 2 & 129 \\
H3DGS $\tau=6$	& \cMark & \fMark & \fMark & \fMark  & \fMark  & \textit{13} & \textit{19} & \textit{5} & \textit{97} \\
3DGS                  & \fMark & 43 & 8 & 38 & 8 & 38 & 8 & 22 & 15 \\
\midrule
Full representation   & \cMark & \fMark & \fMark & 26 & 18 & 29 & 19 & 17 & 31 \\
Ours - single cluster & \cMark & 42 & 6 & 34 & 5 & 41 & 7 & 23 & 10 \\
Ours                  & \cMark & 41 & 7 & 35 & 5 & 43 & 7 & 22 & 9 \\
\bottomrule
\end{tabular}
\vspace{6pt}
\caption{\textbf{Mobile experiment.} We estimate and compare the rendering speed (FPS) and sorting time in milliseconds (S. Time) on various mobile devices. (\fMark) means rendering crashed; HQ=high quality.
\label{tab:mobile-experiment}%
}
\end{table}

\PAR{Rendering on mobile and low-power devices}
Finally, we benchmark rendering speed on four devices: iPhone 13 Mini, iPhone 15 Pro, HP Elite Dragonfly Chromebook, and MacBook Air 13 inch, using the web-based 3DGS renderer by Mark Kellogg~\cite{kellogg}, which performs rendering with asynchronous Gaussian sorting running in parallel with the rasterization.
We report both FPS (rasterization only) and sorting time (S. time) in milliseconds in \Cref{tab:mobile-experiment}.
For H3DGS~\cite{kerbl2024h3dgs}, we show both the full model ($\tau = 0$) and the default ($\tau = 6$) configuration.
Vanilla 3DGS renders fast due to its low Gaussian count but at the cost of poor quality (see \Cref{tab:h3dgs}; PSNR @ 25.42 vs. 26.57 for ours). H3DGS instead fails on iPhones due to memory limitations and cannot render in real time on laptops. Notably, iPhone 13 Mini outperforms the 15 Pro in FPS due to its smaller display.
Our method runs efficiently across all devices and achieves the best performance on laptops.

\section{Conclusion}
We introduced a novel level-of-detail (LOD) approach for 3D Gaussian Splatting that enables real-time rendering of large-scale scenes, even on memory-constrained devices. Our method combines a multi-level LOD representation with chunk-based rendering to avoid per-frame overhead by precomputing active Gaussian sets for spatial regions. We further proposed an automatic threshold selection strategy and a two-cluster opacity blending scheme to ensure smooth transitions between chunks. Extensive experiments on both indoor and outdoor datasets demonstrate that our method outperforms state-of-the-art baselines in both rendering quality and speed. Importantly, our approach is deployable on mobile devices, achieving real-time performance where other methods fail.%

\PAR{Limitations}
While our method enables real-time rendering on mobile devices, it assumes that loading Gaussians—and reloading them when crossing chunk boundaries—can be performed efficiently. In practice, this would require optimized web servers and effective compression protocols to stream Gaussians to the device in real time, which we leave as future work.

\medskip
{
\small
\bibliography{bibliography}
}

\clearpage
\appendix
{
\section{Supplementary Material}
\let\section\subsection
\section{Video}
In the attached video\ifpublic{ (\url{https://lodge-gs.github.io/video.html})}, we present a qualitative comparison between our method and several existing approaches. 
First, we compare our method with ZipNeRF~\cite{barron2023zipnerf}, Octree-GS~\cite{ren2024octree}, H3DGS~\cite{kerbl2024h3dgs}, and 3DGS~\cite{kerbl20233dgs} on the SmallCity and Campus scenes from the Hierarchical 3DGS dataset~\cite{kerbl2024h3dgs}, as well as the NYC and London scenes from the ZipNeRF dataset~\cite{barron2023zipnerf}.
ZipNeRF produces less sharp results, particularly for nearby cars and the ground. Octree-GS exhibits over-exposure and reduced detail in close-up cars. H3DGS achieves comparable visual quality to ours but is significantly slower to render. 3DGS, on the other hand, suffers from blurriness and a noticeable loss of detail.
Next, we demonstrate how our LOD representation effectively reduces the number of visible Gaussians, thereby accelerating the rendering process.
Finally, we show temporal artifacts when using `LOD + chunks' (without opacity blending), and show how opacity blending removes these artifacts.

\section{Implementational Details}
In our method, we first optimize the `base' model -- \ie, a vanilla 3DGS reconstruction—and then build the LOD representation on top. 
Given the `base' model, we build the LOD representation (as described in Section `LOD structure optimization steps'). We then build the LOD chunks as described in Section 3 of the main paper. In Section `Selecting chunk centers' we give details on how the chunks are selected.  To build the `base' 3DGS reconstruction, we combine several improvements from recent literature~\cite{yu2023mip3dgs,kerbl2024h3dgs,niemeyer2024radsplat} which we describe in Section `Detailed on training the full representation'.

\PAR{LOD structure optimization steps} In this section we detail the exact steps and parameters used for creating the LOD gaussians. 
Given a sequence of depth thresholds $d_l$, we build the LOD sets $\mathcal{G}^{(l)}$ from the finest to the coarsest level. We start from the set $\mathcal{G}^{(0)}$, iteratively adding more levels. At step $l$, we construct the set $\mathcal{G}^{(l)}$ via applying a 3D smoothing filter to all Gaussians in $\mathcal{G}^{(0)}$. Note that the smoothing filter is only added to the Gaussian function, whereas the parameters remain unchanged. Therefore, the optimization process cannot decrease Gaussians' sizes below the 3D filter size. Next, we prune all Gaussians in $\mathcal{G}^{(l)}$ using the importance score pruning and then perform $1\,000$ optimization steps. The optimization steps use the LOD rendering procedure (with LOD levels up to $l$) described in the previous section, with the modification that depth $d_l$ is replaced by a random number drawn from a uniform distribution $\mathcal{U}(0.7 d_l, 1.3 d_l)$.
This makes the representation more robust towards cameras lying outside of the training trajectory.
We use the same loss function (DSSIM+L1) as during the standard optimization, however, we only update parameters of the Gaussians in set $G_l$. This process of pruning and fine-tuning is repeated in total three times, with importance score thresholds of $0.2\gamma$, $0.6\gamma$, and $\gamma$, where $\gamma$ is a hyperparameter. Therefore, we need overall $N_{\text{levels}} \cdot 3 \cdot 1\,000$ optimization iterations.

\PAR{Selecting chunk centers}
We split the scene into chunks, by performing a K-means clustering on the training camera positions. We set the number of clusters $N_{\text{clusters}}$ as follows:
\begin{equation}
N_{\text{clusters}} = \frac{4}{d_1} \max_i\| \mathbf{c_i} - \bar{\mathbf{c}}\|\,, \quad\quad\quad \bar{\mathbf{c}} = \frac{1}{N_{c}}\sum_j^{N_c} \mathbf{c}_j \,,
\end{equation}
where $N_c$ is the number of training cameras, $c_i$ are camera positions, and $d_1$ is first LOD depth threshold. For outdoor scenes this roughly corresponds (empirically) to cluster sizes of 5 meters.

\PAR{Details on training the full representation}
In our method, the LOD representation is build from an existing 3DGS representation. To this end, we extended vanilla 3DGS with recent improvements~\cite{yu2023mip3dgs,kerbl2024h3dgs,niemeyer2024radsplat} to achieve better quality of the base representation.
We employ the original 3DGS renderer augmented with a 2D filter, as proposed in Mip-Splatting~\cite{yu2023mip3dgs}. However, for densification, we adopt the modified strategy from H3DGS~\cite{kerbl2024h3dgs}. Specifically, we replace the original hard threshold on the 2D position gradient norm used for cloning/splitting with the condition:
\begin{equation}
\text{grad\_position\_2D} \cdot \text{max\_radii\_2D} \cdot (\text{opacity})^{1/5} >= \text{threshold} \,,
\end{equation}
where max\_radii\_2D is the largest radius the Gaussian projects into (since the last densification).
Moreover, instead of averaging the gradient statistics, we take the maximum. 
We use a gradient threshold of $0.015$ for the Hierarchical 3DGS~\cite{kerbl2024h3dgs} dataset and $0.03$ for the ZipNeRF~\cite{barron2023zipnerf} dataset.
All models were trained for for $30\,000$ iterations. We reset opacity every $3\,000$ steps until iteration $15\,000$ and apply densification every $300$ steps from step $600$ to $15\,000$. Importance score pruning is performed at steps $8\,000$, $16\,000$, and $24\,000$ with threshold of $0.02$ for all scenes, except Campus, where we use lower threshold of $0.01$.

\section{Datasets}
To validate our method, we use two larger-scale datasets: two outdoor scenes used in the Hierarchical 3DGS paper \cite{kerbl2024h3dgs} and three indoor scenes from the Zip-NeRF paper \cite{barron2023zipnerf}.

\PAR{Hierarchical 3DGS dataset~\cite{kerbl2024h3dgs}}. The public release of the dataset contains two urban scenes (SmallCity and Campus). To collect the dataset, authors used a bicycle helmet on which they mounted 6 GoPro HERO6 Black cameras (5 for the Campus scene). The SmallCity scene was collected by riding a bicycle at around $6 - 7$km/h, while Campus was captured on foot wearing the helmet. Poses were estimated using COLMAP with custom parameters and hierarchical mapper. Additinal per-chunk bundle adjustment was performed, see Hierarchical 3DGS~\cite{kerbl2024h3dgs}. In our experiments, we use the official train/test split provided by the authors. We also use the official segmentation masks provided with the dataset to remove license plates, pedestrians, and moving cars.

\PAR{Zip-NeRF dataset~\cite{barron2023zipnerf}}.
The dataset contains four large-scale indoor scenes scenes: Berlin, Alameda, London, and NYC, (1000-2000 photos each) captured using fisheye cameras. 
We use the provided undistorted images and following Zip-NeRF~\cite{barron2023zipnerf}, we use downscale factor of two resulting in downsampled resolutions between 1392 × 793 and 2000 × 1140 depending on scene. Note, that we use the provided downscaled JPEG images for training and evaluation.
In our experiments, we only use Alameda, London, and NYC, because for some baselines the Berlin scene cannot be fitted into GPU memory. Same as Zip-NeRF, we take each 8$^\text{th}$ image as testing image (when sorted alphabetically).

\section{Evaluation Protocol \& Baselines}
We report standard PSNR, SSIM, and LPIPS metrics. To ensure a fair evaluation and reproducibility, before computing metrics, we round the predictions to \texttt{uint8} range. For the Hierarchical 3DGS dataset, we mask the prediction and the ground truth image by the provided mask - replacing the pixels inside the mask with black color. For SSIM, we use commonly used default values and average across pixels and color channels. For LPIPS~\cite{zhang2018lpips}, we use the VGG network (unlike some baselines which used AlexNet with lower values).
To compute FPS, we compute rendering times for each testing image. We compute each per-image rendering time $7$ times and take a median to ensure precise measurement. The resulting FPS is then one over average rendering time for all test images. In the rendering time, we only include the rendering part, not the time required to move the data from GPU memory at the end of the rendering.
When reporting the number of Gaussians $\#G$, we report the average number of Gaussian loaded in GPU memory when rendering. For methods which constructs LOD splits on the fly this is the total number of Gaussians in all LOD sets (if applicable). Note, that for Scaffold-GS~\cite{lu2023scaffoldgs} and Octree-GS~\cite{ren2024octree}, some attributes (colors) are not stored explicitly, but are decoded on-the-fly lowering memory requirements.
All baselines and our method were trained and evaluated on a single NVIDIA A100 SXM4 40 GB GPU.

\PAR{Baselines}
For the baselines, we include the following:
\textbf{Zip-NeRF}~\cite{barron2023zipnerf} which is the current state-of-the-art on the Zip-NeRF dataset. Note, that this method was trained for 200K iterations - much longer than other baselines and does not achieve real-time rendering. We only include it for reference.
\textbf{RadSplat}~\cite{niemeyer2024radsplat} uses Zip-NeRF for initialization and is, therefore, much more expensive to train than other baselines. It uses the same strategy as our method to remove less important Gaussians during training. The method used Zip-NeRF checkpoint trained for 200K and performed additional 45K training iterations to optimize 3DGS representation.
\textbf{3DGS}~\cite{kerbl20233dgs} is the basis for most other methods. We use the default parameters, training for 30K iterations.
\textbf{Mip-Splatting}~\cite{yu2023mip3dgs} augments 3DGS with antialiasing 2D and 3D filters. In our method, we also use the same 2D filter and we use the 3D filter when building LOD levels.
\textbf{Scaffold-GS}~\cite{lu2023scaffoldgs} reduces memory requirements by decoding attributes of Gaussians on the fly from MLP. Therefore it is important baseline for rendering speed as our method stores all attributes explicitly.
\textbf{H3DGS}~\cite{kerbl2024h3dgs} starts from a 3DGS representation optimized at the finest level (same as ours) and builds a representation by iteratively merging Gaussians, however, at render time all Gaussians need to be kept in memory and LOD split needs to be built with every rendering call. The method is currently SoTA for the Hierarchical 3DGS dataset. The
\textbf{FLOD}~\cite{seo2024flod}, \textbf{CityGS}~\cite{liu2024citygs}, and \textbf{Octree-GS}~\cite{ren2024octree} are both recent LOD 3DGS-based approaches. Octree-GS was trained for 40K iterations, while FLOD used 100K iterations for the 5 LOD levels. At rendering, we use the selective 3, 4, 5 rendering split. For all baselines we modified the code to accept masks (masking loss gradient propagation in masked regions). We have performed a limited hyperparameter search to get best performance for the datasets.

\section{Broader Impact}
We are not aware of any potential misuse of our method. On the contrary, we believe it enables positive applications, particularly by making large-scale 3D reconstructions deployable on mobile and resource-constrained devices. This can enhance navigation in complex urban environments, support AR experiences, and improve accessibility for users with visual impairments. Our method may also benefit fields like robotics, autonomous driving, and cultural heritage preservation by providing efficient scene representations.
}

\immediate\closein\imgstream

\end{document}